\documentclass[letterpaper, 10 pt, conference]{ieeeconf}  

\IEEEoverridecommandlockouts                              

\usepackage{graphics} 
\usepackage{epsfig} 
\usepackage{mathptmx} 
\usepackage{times} 
\usepackage[ruled, vlined, linesnumbered, commentsnumbered, longend]{algorithm2e}
\usepackage{booktabs}
\usepackage{pifont}
\usepackage{tabularx}
\usepackage{multicol}
\usepackage{color}
\usepackage{balance}
\usepackage{multirow}
\usepackage[normalem]{ulem}
\usepackage{amsmath} 
\usepackage{amssymb}  
\useunder{\uline}{\ul}{}
\usepackage{url}

\newcommand{\que}[1]{{\color{black} #1}}
\newcommand{\xf}[1]{{\color{black} #1}}
\newcommand{\shs}[1]{{\color{black} #1}}
\newcommand{\wt}[1]{{\color{black} #1}}

\title{\LARGE \bf
Balanced Line Coverage in Large-scale Urban Scene
}

\author{Hangsong Su$^{1,\dag}$, Feng Xue$^{1,\dag}$, Runze Guo$^{1}$, and Anlong Ming$^{1,*}$
	\thanks{\dag Equal Contribution}
	\thanks{*Corresponding Author}
	\thanks{$^{1}$Beijing University of Posts and Telecommunications, Beijing, China,
		{\tt\small \{suhangsong,xuefeng,grz,mal\}@bupt.edu.cn}}%
}

\begin{document}

\maketitle
\thispagestyle{empty}
\pagestyle{empty}

\begin{abstract}
\xf{Line coverage is to cover linear infrastructure modeled as 1D segments by robots,
which received attention in recent years.
With the increasing urbanization,
the area of \wt{the} city and the density of infrastructure continues to increase,
which brings two issues:
(1) Due to the energy constraint,
it is hard for the homogeneous robot team to cover the large-scale linear infrastructure \que{starting} from one depot;
(2) In the large urban scene,
the imbalance of robots' path greatly extends the time cost of \wt{the} multi-robot system,
which is more serious than that in smaller-size scenes.
To address these issues,
we propose a heterogeneous multi-robot approach consisting of several teams,
each of which contains one transportation robot (TRob) and \que{several coverage robots (CRobs)}.
Firstly,
a balanced graph partitioning (BGP) algorithm is proposed to divide the road network into several similar-size sub-graphs,
and then the TRob delivers a group of CRobs to the sub-graph region quickly.
Secondly,
a balanced ulusoy partitioning (BUP) algorithm is proposed to extract similar-length tours for each CRob from the sub-graph.
Abundant experiments are conducted on seven road networks ranging in scales that are collected in this paper.
Our method achieves robot utilization of 90\% and the best maximal tour length at the cost of a small increase in total tour length,
which further minimizes the time cost of \wt{the} whole system.
The source code and the road networks are available at \textit{https://github.com/suhangsong/BLC-LargeScale}.
}

\end{abstract}

\section{Introduction}
\begin{figure}[t]
    \centering
    \includegraphics[width=\columnwidth]{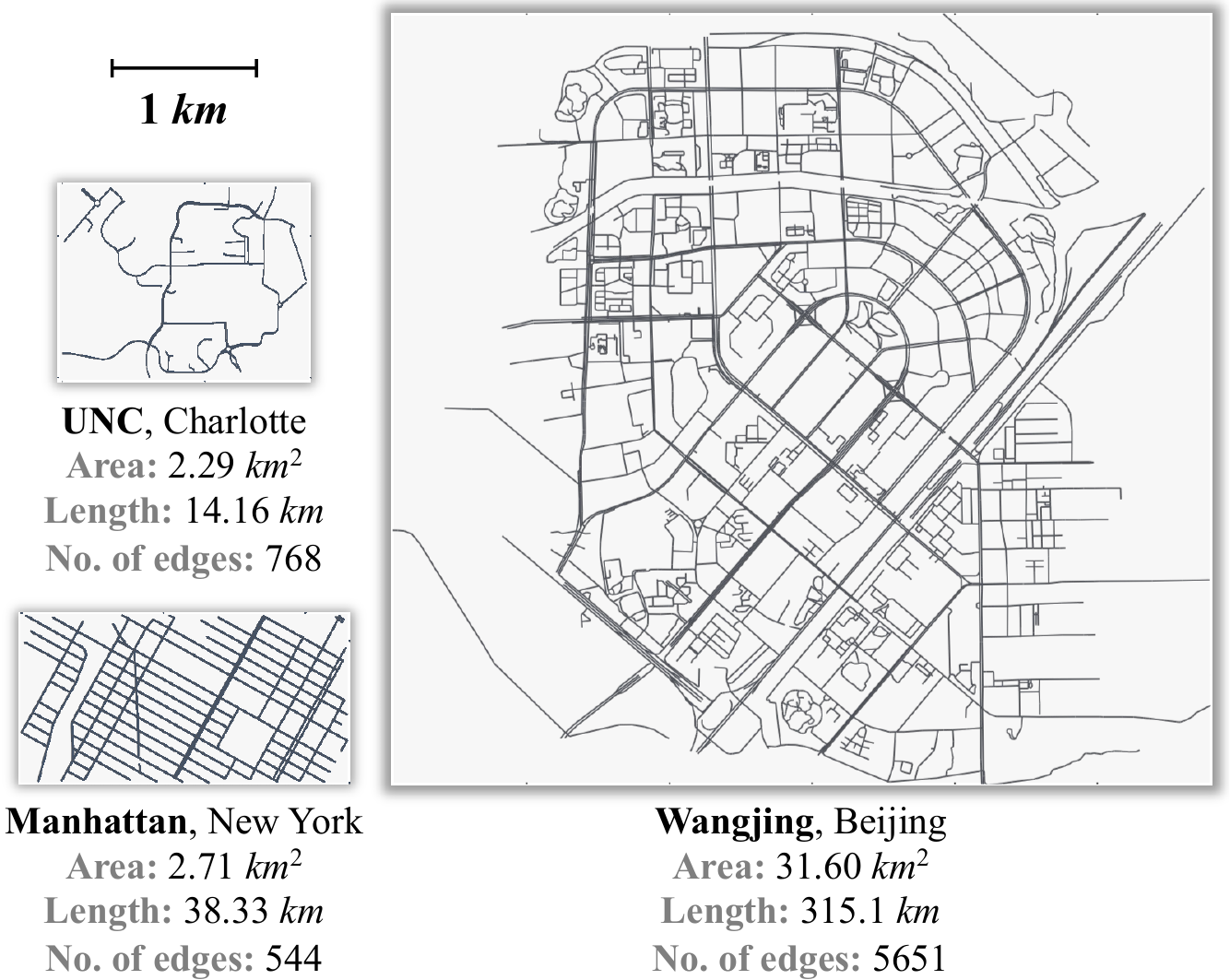}
    \caption{Visualization comparison between the road networks used in previous method \cite{SauravAgarwal2020LineCW} and the medium-sized one captured in this paper.
    }
    \label{fig:map1}
    \vspace{-7pt}
\end{figure}

\xf{Multi-robot systems are widely applied in inspection \cite{9476813,9560826,9384182},
monitoring \cite{MatthewDunbabin2012RobotsFE,9636854,9653829},
search-rescue operations \cite{YoonchangSung2017AlgorithmFS}.
Generally,
these tasks can be performed by covering the linear environment features such as road networks, gas lines, power lines, and etc,
i.e., the multi-robot line coverage.
The prior line coverage algorithms \cite{LingXu2011AnEA,SauravAgarwal2020LineCW} have achieved outstanding performance in \wt{the} city with limited \wt{areas} or sparse road networks,
as shown in two images on the left side of Fig. \ref{fig:map1}.

However,
with the increasing area and density of infrastructure (see the right image in Fig. \ref{fig:map1}),
there are two inevitable issues \que{arisen}.
Firstly, it is difficult for a group of energy-constrained unmanned aerial vehicles (UAVs) or unmanned ground vehicles (UGVs) to conduct line coverage from the same depot.
Secondly,
the existing line coverage methods fail to balance the robots' workloads in an available computation time,
thus increase the time cost of the whole system,
which is stark in a large-scale road network.
}


\xf{In this paper,
we introduce a heterogeneous multi-robot method to achieve line coverage in \wt{a} large-scale road network that is represented by an undirected graph.
It consists of multiple robotic teams,
each containing a transportation robot (TRob) and \que{several coverage robots (CRobs)}.
The pipeline of line coverage is divided into two stages,
namely,
firstly partitioning this graph,
and then performing line coverage on sub-graphs by a robot team individually.
In the first stage,
to achieve \que{the} similar workloads of various robot teams,
a balanced graph partitioning (BGP) algorithm is proposed to obtain several similar-size sub-graphs that \wt{need} to be covered.
In the second stage,
we design a balanced ulusoy partitioning (BUP) algorithm,
a modified version of ulusoy partitioning algorithm \cite{GndzUlusoy1985TheFS,OsmanParlaktuna2009MultirobotSC}.
It iteratively shrinks the length range of the available tour to reduce the difference in length between coverage \que{tours} of robots.
To verify the effectiveness of the proposed method,
we collect multiple road networks that vary greatly in scale.
\wt{Through} the experiments on the collected data,
we verify the effectiveness of the BGP algorithm and the BUP algorithm in the two stages proposed in this paper.
In addition, we evaluate the whole proposed method by \wt{simulation}.

}



\begin{figure*}[t]
    \centering
    \includegraphics[width=\textwidth]{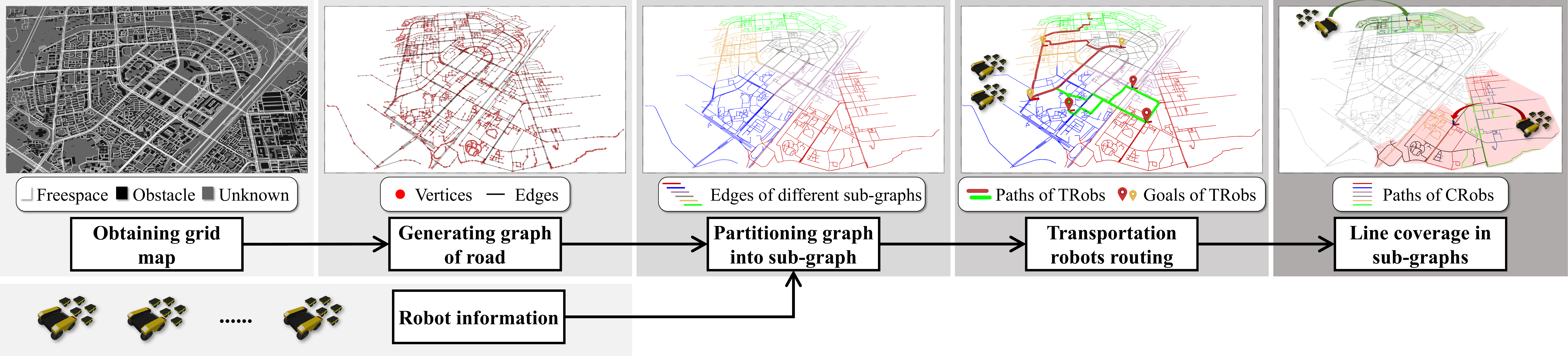}
    \caption{
        The pipeline of the heterogeneous multi-robot line coverage approach.
        For the map captured by Tencent Location Services,
        we input the grid map.
        For the map captured by OpenStreetMap,
        we input the graph of road networks directly.
    }
    \label{fig:overview}
    \vspace{-4pt}
\end{figure*}

\xf{
The main contributions in this paper are listed as follows:
\begin{itemize}
    \item
    To cover the large-scale linear feature,
    we propose a heterogeneous multi-robot approach.
    Abundant experiments on multiple road networks demonstrate the effectiveness of our method.

    
    \item For graph partitioning,
    the balanced graph partitioning algorithm (BGP),
    i.e., a modified version of k-medoids clustering algorithm,
    is proposed to greatly improve the length balance between sub-graphs.
    
    
    \item For the stage of line coverage,
    we propose the balanced Ulusoy partitioning algorithm (BUP) to obtain balanced tours for robots,
    making our method to cover a large-scale road network in less time than others.
    
\end{itemize}
}

\section{Related Work}

\subsection{Heterogeneous robot system}
Heterogeneous robot systems have been applied in various tasks,
such as search and rescue \cite{9384153}, path planning \cite{9381577}, and localization \cite{4059130}.
For the coverage problem,
heterogeneous robot systems are also applied in area coverage.
Yu \textit{et al.} \cite{KevinYu2019CoverageOA} propose to use UGV to charge and transport UAV,
reducing the flight time of the UAV covering large areas. 
Deng \textit{et al.} \cite{DiDeng2019ConstrainedHV} propose a large-area-coverage planning framework to optimize the paths for UAVs and UGVs. 
The area that needs to be covered is divided into multiple sub-regions that are assigned to a fleet of UAVs,
and simultaneously optimize the trails of UAVs and UGVs.
\xf{Different from the works using multiple heterogeneous robots to conduct the area coverage and 
inspired by them,
this paper propose to cover large-scale line feature by using multiple heterogeneous robots.}


\subsection{Line coverage with multiple robots}
According to whether the energy of the robot is considered, the line coverage problem is divided into two categories. 
For the robots with unlimited energy,
Xu \textit{et al.} \cite{LingXu2011AnEA} propose to improve the performance of line coverage by boosting balance of paths on incomplete prior map.
For the robots with limited energy, 
Parlaktuna \textit{et al.} \cite{OsmanParlaktuna2009MultirobotSC} and Sipahioglu \textit{et al.} \cite{AydinSipahioglu2010EnergyCM} 
formulate the problem as Capacitated Arc Routing Problem with Deadheading Demands (CARPDD) \cite{GokhanKirlik2012CapacitatedAR},
a variant of Capacitated Arc Routing Problem(CARP), 
and solved the problem by modifying the ulusoy partitioning algorithm. 
Agarwal \textit{et al.} \cite{SauravAgarwal2020LineCW} study the case in which costs and demands are completely asymmetric,
and propose a fast heuristic algorithm to solve it. 
In addition, the exact algorithm proposed by \cite{GokhanKirlik2012CapacitatedAR,EnricoBartolini2013AnEA} can solve CARPDD precisely on a small-scale road network.
\shs{Considering the robots with limited energy,}
this paper focus on conducting balanced line coverage in large-scale urban scene.



\subsection{Task balance of multiple robots}
Task balance between robots has been studied in several tasks,
such as vehicle routing problem \cite{applegate2002solution},
area coverage \cite{wang2021balanced}, etc. 
For the area coverage,
Kapoutsis \textit{et al.} \cite{AthanasiosChKapoutsis2017DARPDA} improve the workload balance among robots 
with a balanced partition map. 
By improving the balance of robot workloads, 
Collins \textit{et al.} \cite{LeightonCollins2021ScalableCP} and Tang \textit{et al.} \cite{tang2021mstc} achieve the shortest completion time. 
For the algorithm solving CARP,
the memetic algorithm (MA) is used to optimize the balance of \wt{the} path 
\cite{YiMei2011DecompositionBasedMA,MuhilanRamamoorthy2021MAABCAM}.
However,
MA \wt{achieves} excellent results at the cost of more computation time.
For the problem of multi-robot line coverage with limited energy, 
there are few studies on \wt{the} task balance of robots.
To reduce the completion time of line coverage,
we balance the areas of sub-graphs and the workloads of robots. 


\section{Overview}

\xf{
Given the grid map of an urban road network,
which is commonly used in robot navigation,
we firstly employ \wt{the} generalized voronoi diagram (GVD) \cite{ErcanUAcar2006SensorbasedCW} to obtain a graph of the road $G = (V,E)$,
where $V$ is the set of vertices,
and $E$ is the set of edges.
The goal of the multi-robot line coverage is to plan each robot's path to completely travel through all edges in $E$ with the shortest possible time,
as shown in Fig.\ref{fig:overview}.
However, 
such immense and miscellaneous roads greatly increase the scale and complexity of graph $G$,
making energy-constrained robots hard to cover all edges when all \que{robots} start and end their tours at \wt{the} same depot.
For this reason,
we design a heterogeneous multi-robot approach consisting of several robot teams,
each of which contains a transportation robot (TRob) and several coverage robots (CRob).
TRobs provide the charging service and transport the CRobs to a sub-graph $G^{'}_{i}=(V^{'}_{i},E^{'}_{i})$,
where $V^{'}_i\subset V$ and
$E^{'}_i\subset E$.
\wt{CRobs of \wt{the} same team cover the sub-graph $G^{'}_i$.}
To be specific,
our approach is divided into two stages:


\begin{enumerate}
    \item \textbf{Balanced graph partitioning and assignment}:
    Given the graph $G$,
    this stage divides $G$ into $k$ sub-graphs $G^{'}_i$ with \que{balanced lengths},
    where $G = G^{'}_1\cup G^{'}_2\cup...\cup G^{'}_k$.
    For each sub-graph $G^{'}_i$,
    the vertex with the smallest sum of path length to other vertices in $G^{'}_i$ is taken as a goal point of TRob.
    Each TRob obtains a set of goal \que{points} that need to be traversed.
    
    
    
    

    \item \textbf{Balanced line coverage in sub-graphs}: 
    When a TRob reaches the goal point inside $G^{'}_i$,
    the balanced ulusoy partitioning (BUP) algorithm is designed to plan the balanced tour for each CRobs to traverse all edges $E^{'}_i$.
    
\end{enumerate}
}

\section{Balanced Graph Partitioning and Assignment}
\subsection{Balanced graph partitioning}
\xf{The prior works \cite{LingXu2011AnEA,SauravAgarwal2020LineCW} consider \wt{graph} partitioning as a clustering problem.
However, 
\shs{they neglect to equalize the lengths of the clusters (i.e., the sub-graphs).}
Thus, the \que{workload} of the robot teams vary greatly,
leading to inefficiency of the whole multi-robot system.
To obtain the balanced sub-graphs,
we propose the balanced graph partitioning (BGP) algorithm,
as given in Algorithm \ref{alg:graphpartition},
which is composed of three parts:
(1) K-Medoids clustering with dynamic scale factor,
(2) disconnected sub-graphs elimination,
(3) boundary edge re-assignment.}
\wt{Note that,
the notation for set in Algorithm \ref{alg:graphpartition} is different from that in text.
For example,
$\{G^{'}_1,G^{'}_2,...,G^{'}_k\}$ is expressed as $G^{'}_{1...k}$.}
\xf{Before partitioning the graph $G=(V,E)$,
\shs{we calculate the number of sub-graphs by the maximal length that a team can cover without recharging:}
\begin{equation}
\label{eq:clusternumber}
k = \left \lceil \sum\nolimits_{e\in E}{|e|} / (\alpha MQ) \right \rceil
\end{equation}
where $Q$ is the maximum driving distance of CRob,
$M$ is the number of CRobs in a team,
$|e|$ is the length of edge $e$.
$\alpha\in(0,1)$ is a constant to adjust the scale of sub-graphs,
which is set empirically.
}

\xf{
\subsubsection{K-Medoids clustering with dynamic scale factor}
This part aims to cluster graph $G$ into $k$ sub-graphs.
Referring to \cite{TeofiloFGonzalez1985ClusteringTM},
\shs{initially the first vertex in the graph is taken as the first cluster's centroid $c_1$, followed by $k-1$ iterations.}
In each iteration,
the vertex with the largest 
\shs{length} 
of \que{the} shortest path to the determined centroids is taken as a new cluster's centroid,
which is terminated until gaining all centroids $\{c_1,c_2,...,c_k\}$.

After the initialization,
we assign each edge in $E$ to its nearest cluster and recompute the centroids of all clusters,
which is iterated until the length ratio between \que{the} largest sub-graph and the smallest sub-graph is less than $\varepsilon$ or the loop count is larger than $\tau_1$.
The detailed iterative process is given as follows.
Firstly,
the distances from each edge to all centroids $\{c_1,c_2,...,c_k\}$ are calculated as follows:
\begin{equation}
\label{eq:distancecalculation}
    d_{ij} = s_i\min(\textbf{d}(v_a, c_i),\textbf{d}(v_b, c_i)), \,\,\, e_j = (v_a,v_b)\in E , \,\, i\in[1,k]
\end{equation}
where $\textbf{d}(,)$ is the length of \wt{the} shortest path between two vertices,
\shs{$e_j$ for the edge between vertex $v_a$ and vertex $v_b$,}
$d_{ij}$ for the shortest distance from $i$-th centroid to $j$-th edge.
Note that,
$s_i$ is the dynamic scale factor of centroid $c_i$,
which is initialized to $1$.
Secondly,
each edge $e_j$ and its two vertices are assigned to its nearest cluster:
$E^{'}_{\mathbf{x}}=E^{'}_{\mathbf{x}}\cup\{e_j\}$,
$V^{'}_{\mathbf{x}}=V^{'}_{\mathbf{x}}\cup\{v_p,v_q\}$,
where $\mathbf{x} = \mathop{\arg\min}\nolimits_{i}(d_{ij}), \,i\in[1,k]$ is the index of the nearest cluster of edge $e_j$.
In this way,
the sub-graphs $\{G^{'}_1,G^{'}_2,...,G^{'}_k\}$ can be obtained,
where $G^{'}_i=\{E^{'}_{i},V^{'}_{i}\}$.
To reserve the optimal sub-graphs result,
these generated sub-graphs are saved as the best partitioning result if the length ratio between the largest sub-graph and the smallest one is smaller than that of the prior best result.
In addition,
the centroid of each sub-graph is re-assigned as its central vertex.
Finally, \que{inspired by \cite{AthanasiosChKapoutsis2017DARPDA}}, 
we update the scale factor of each cluster $G^{'}_i$ by using the difference 
\shs{between the length of $G^{'}_i$ and the average length of all sub-graphs:}
\shs{
\begin{equation}
\begin{split}
\label{eq:dynamicscalefactor}
s_i &= s_i + \eta_1 \big((l_i - l_{avg})/l_{avg}\big) + \eta_2 \big((l_i - l_{avg})/l_{avg}\big) ^ 3\\
l_i &= \sum\nolimits_{e_j\in E^{'}_i}|e_j|, \,\, l_{avg} = \frac{1}{k} \sum\nolimits_{E^{'}\in\{E^{'}_\mathbf{i}|\mathbf{i}\in[1,k] \}}\sum\nolimits_{e_j\in E^{'}}|e_j|
\end{split}
\end{equation}}
where $\eta_1$ and $\eta_2$ denote the positive coefficient.

\begin{algorithm}[!t]
\small
    \caption{Balanced Graph Partitioning}
    \label{alg:graphpartition}
    
    \SetKwInOut{KwIn}{Input}
    \SetKwInOut{KwOut}{Output}

    \KwIn{ $G(V,E)$, number of sub-graphs $k$}
    \KwOut{ $k$ sub-graphs $G^{'}_{1 \dots k}$}
    
    $c_{1\dots k} \gets$ InitClusterCentroids($G$);
    
    $s_{1 \dots k} \gets 1$;
    
    \tcc{K-medoids with scale factor}
    $loop \gets 0$; 
    $ratio^{best}, ratio \gets \infty $;
    
    \SetAlgoVlined
    \While{$loop < \tau_1$ {\bf and} $ratio > \varepsilon $}{
        $G^{'}_{1 \dots k} \gets $ ClusterEdges($G, c_{1 \dots k}, s_{1 \dots k}$);
        
        $ratio\gets$MaxLength$(G^{'}_{1 \dots k})/$MinLength$(G^{'}_{1 \dots k})$;

        \SetAlgoVlined
        \If{$ratio < ratio^{best}$}{
            $ratio^{best} \gets ratio$; 
            $G^{best}_{1 \dots k} \gets G^{'}_{1 \dots k}$;
        
            $c_{1 \dots k} \gets $RecomputeCentroids($G^{'}_{1 \dots k} $);
        }
        $s_{1 \dots k} \gets $UpdateScaleFactor($G^{'}_{1 \dots k}, s_{1 \dots k}$);
        
        $loop = loop + 1$;
    }

    \tcc{Disconnected sub-graph elimination}
    $\{G_i^{isolated}\}\gets$GetDisconnectGraph($G^{best}_{1 \dots k}$);
    
    $G^{'}_{1 \dots k}, G^{best}_{1 \dots k} \gets$AssignToAdjacent($G^{best}_{1 \dots k},G_i^{isolated}$);
    
    $ratio^{best},ratio\gets$MaxLength$(G^{'}_{1 \dots k})/$MinLength$(G^{'}_{1 \dots k})$;
    
    \tcc{Boundary edge re-assignment}
        $loop \gets 0$;
        
        \While{$loop < \tau_2$ {\bf and} $ratio > \varepsilon $}{
            $E^{boundary}\gets$GetBoundaryEdges($G^{'}_{1 \dots k}$);
        
            $G^{'}_{1 \dots k}\gets$GreedyAllocation($G^{'}_{1 \dots k},E^{boundary}$);
        
            $ratio \gets$MaxLength$(G^{'}_{1 \dots k})/$MinLength$(G^{'}_{1 \dots k}) $;  
            
            \If{$ratio < ratio^{best}$}{
                $ratio^{best} \gets ratio$; 
                $G^{best}_{1 \dots k} \gets G^{'}_{1 \dots k}$;
            }
            $loop = loop+1$;
        }
        $G^{'}_{1 \dots k} \gets G^{best}_{1 \dots k} $;

\end{algorithm}

Referring to Eq.\ref{eq:dynamicscalefactor},
in each iteration,
the sub-graph $G^{'}_i$ with \wt{a} longer path \wt{obtains} a larger scale factor $s_i$,
leading to the distances of all edges to $G^{'}_i$ to be boosted in the next iteration,
according to Eq. \ref{eq:distancecalculation}.
In this way,
the balanced sub-graphs are gradually achieved,
which is denoted as $\{G^{best}_i|i\in[1,k]\}$.
}

\subsubsection{Disconnected sub-graph elimination}
\xf{Although the first part \que{obtains} the balanced sub-graphs $\{G^{best}_i|i\in[1,k]\}$,
it ignores the connectivity of each single sub-graph $G^{best}_i$. 
\shs{Disconnection of sub-graph enlarges the complexity of path planning and the length of the paths.}
To eliminate the disconnected sub-graph,
we extract the set of connected edges from a sub-graph.
If the set number is greater than 1,
the sets with 
\shs{non-maximum lengths} are assigned to their adjacent
sub-graph, respectively,
otherwise no operation is performed.
Since the isolated edges are merged into the adjacent sub-graph,
each sub-graph is adjusted into a simply connected graph.
However,
this process results in the imbalanced \que{edge} lengths of sub-graphs.}

\subsubsection{Fine-tune by boundary edge re-assignment}
\xf{To ensure the balance of sub-graphs,
we fine-tune the sub-graphs by re-assigning the edge at the boundary of each sub-graph.
Each boundary edge is assigned to its adjacent smallest sub-graph,
which is iterated until the length ratio between the largest sub-graph and the smallest sub-graph is less than $\varepsilon$ or the loop count is larger than a constant $\tau_2$.}


\subsection{Transportation Robots Routing}
After partitioning graph $G$,
all sub-graphs $\{G^{'}_1,G^{'}_2,...,G^{'}_k\}$ are assigned to the robot teams.
\xf{To minimize the tour length of the TRobs,
the task allocation of TRobs is taken as a multiple traveling salesman Problem (MTSP) \cite{gutin2006traveling}.
And the genetic algorithm \cite{yuan2013new} is directly employed to allocate an order of sub-graph centroid to each TRob for point coverage.}

\section{Balanced Line Coverage in Sub-Graphs}
\xf{In this section,
we propose a balanced ulusoy partitioning (BUP) algorithm.
It aims to find a set of tours on the sub-graph $G^{'}_i$ that minimizes the difference in length and the total cost of \wt{travel}.
In graph $G^{'}_i=(V^{'}_i, E^{'}_i)$,
all edges $E^{'}_i$ are the linear features to be covered, i.e., the required edge.
And other edges outside this sub-graph, i.e., $E-E^{'}_i$, are the non-required edge used to travel from one vertex to another.}
The detailed steps \wt{are} given in Algorithm \ref{alg:BalancedUlusoy}.

\subsection{Ulusoy Partitioning}
\xf{Since our algorithm is a modified version of ulusoy partitioning (UP) algorithm \cite{OsmanParlaktuna2009MultirobotSC},
we briefly review \wt{the} UP algorithm and the related symbols in this subsection.

The UP algorithm consists of two steps.
Firstly, it constructs the Euler circuit on graph $G^{'}_i$,
represented by a directed graph $D(V,A)$,
where $V=\{v_0, v_1, \dots, v_r\}$ is the vertices set,
each vertex $v_j(0 \leq j \leq r)$ correspond a vertex in sub-graph $G_i^{'}$,
and $v_0=c_i$ is the depot of this sub-graph.
$A=\{(\protect\overrightarrow{v_a, v_b})\mid 0 \leq a < b \leq r\}$ is the arcs set, 
and each arc $(\protect\overrightarrow{v_a, v_b})$ represents a coverage tour $T_{ab}$ in the sub-graph $G_i^{'}$.
For example,
the tour $T_{ab}$ \que{includes} the edges $\{(v_a, v_{a+1}), (v_{a+1}, v_{a+2}), \dots, (v_{b-1}, v_b)\}$, 
starting and ending at $v_0$.
The length of the arc $(\protect\overrightarrow{v_a, v_b})$ is equal to that of tour $T_{ab}$,
and is less than the energy of CRob.
In the UP algorithm,
the directed graph $D(V,A)$ records all possible circuit partitioning results on the Euler circuit that satisfy the robot \que{energy} constraints.
Secondly,
UP algorithm figure a set of arcs from $v_0$ to $v_r$ whose sum of lengths is the \que{shortest} in all possible results.
\shs{The tour represented by the arc in this set can be assigned to a robot in the team.}

However,
\wt{the} UP algorithm \que{\cite{OsmanParlaktuna2009MultirobotSC}} cannot calculate the coverage paths for a given number of robots.
In addition,
it fails to achieve balanced coverage tours for each \que{robot}.
As a result,
there are usually some robots idling while others are busy working during line coverage.
}

\begin{algorithm}[t]
\small
    \SetKwInOut{KwIn}{Input}
    \SetKwInOut{KwOut}{Output}
    
    \caption{Balanced Ulusoy Partitioning}
    \label{alg:BalancedUlusoy}
    
    \KwIn{$G_{i}^{'}(V_{i}^{'}, E_{i}^{'})$, $G(V,E)$, energy $Q$, CRob number $M$}
    \KwOut{Feasible robots tour $T_{1 \dots t}$}

    \tcc{Obtain a directed graph via \cite{OsmanParlaktuna2009MultirobotSC}}
    $D(V, A) \gets $ UlusoyPartitioning($G_i^{'}, G, Q$);
    
    $V = \left \{ v_0, v_1, \dots, v_r \right \}$; 
    $A = \{(v_a, v_b)\mid 0\leq a < b \leq r\}$;
 
    \tcc{Determine the number of tours}
    $P_s \gets $ GetShortestPath($v_0, v_r$);
    
    $t^{'} \gets $ GetEdgesNumber($P_s$);
    
    $t \gets \left \lceil t^{'}/M \right \rceil \times M $;

    \tcc{Progressively get shortest path}

    $need\_optimize \gets true$;
    $l_{lower} \gets 0$; 
    $l_{upper} \gets Q$;
    
    \While{$need\_optimize$}{

        \ForEach{$(\protect\overrightarrow{v_a, v_b}) \in A$}{
            \lIf{$|\protect\overrightarrow{v_a, v_b}| < l_{lower}$ {\bf or} $|\protect\overrightarrow{v_a, v_b}| > l_{upper}$}{
                delete $(\protect\overrightarrow{v_a, v_b})$
            }
        }
        $need\_optmize, P \gets$ DP($D, t$);
        
        \If{$need\_optimize$}{
            \ForEach{$(\protect\overrightarrow{v_a, v_b}) \in P$ }{
                $T_j \gets $ GetCorrespondTour($(\protect\overrightarrow{v_a, v_b})$);
            }
            $l_{lower} \gets$ MinLength($T_{1, \dots, t}$);
            
            $l_{upper} \gets \beta \times$ MaxLength($T_{1, \dots, t}$);
        }
    }
\end{algorithm}

\subsection{Balanced Ulusoy Partitioning}
\xf{To address the imbalance of coverage tours,
we propose to reduce the length difference between coverage tours in an iterative manner.
To be specific,
given the directed graph $D(V,A)$ captured by UP algorithm \que{\cite{OsmanParlaktuna2009MultirobotSC,VladimirKolmogorov2009BlossomVA}},
we first calculate the expected number of tours 
\shs{for each team: }
\begin{equation}
t = \lceil {t^{'}}/{M} \rceil \times M
\end{equation}
where $M$ is the number of CRob in a team.
\shs{$t^{'}$ is the arc number of the shortest path from $v_0$ to $v_r$ in directed graph $D$, each arc corresponds \wt{to} a tour assigned to the CRob.
$t$ is the expected number of tour for each team. }
In this way,
each CRob obtains an equal number of tours,
and each tour can only be assigned to one CRob.

Subsequently,
we design an iterative process of converging arc lengths to solve $t$ tours.
In each iteration,
the length range of arcs in $A$ is stretched narrower,
that is, removing arcs in $A$ that are less than $l_{lower}$ and greater than $l_{upper}$,
where $[l_{lower},l_{upper}]$ is the length range of remaining arcs in $A$,
initialized to $[0,Q]$.
Then,
a dynamic programming method is designed to calculate the shortest path $P$ when $t$ tours are expected.
In more detail,
a state transition equation is designed as below:
\begin{equation}
    \label{eq:dynamicprogramming}
    dp(\mathbf{v},\mathbf{n})= \min \{ |\protect\overrightarrow{v_k, v_\mathbf{v}}| + dp(k, \mathbf{n} - 1) | {0<k<\mathbf{v}, (\protect\overrightarrow{v_k, v_\mathbf{v}}) \in A} \} 
\end{equation}
where $\mathbf{v}$ is a vertex index in $D$.
$\mathbf{n}$ is the number of arcs in the shortest path from $v_0$ to $v_\mathbf{v}$ in $D$.
$|\protect\overrightarrow{v_k, v_\mathbf{v}}|$ is the length of arc $(\protect\overrightarrow{v_k, v_\mathbf{v}})$.
$dp(\mathbf{v},\mathbf{n})$ is the length of \wt{the} shortest path from $v_0$ to $v_\mathbf{v}$ in $D$ with $\mathbf{n}$ arcs expected.
By using \wt{equation} \ref{eq:dynamicprogramming},
we calculate the shortest path $P$ from $v_0$ to $v_r$ with $t$ arcs expected, i.e., $dp(r,t)$,
where each element in $P$ is an arc in $D$.
If a path consisting of $t$ arcs cannot be found from the remaining arc set $A$,
the path solved in the last iteration is already the optimal result.
Otherwise,
the length range of arcs $[l_{lower},l_{upper}]$ is shrunk to be narrower.
$l_{lower}$ is updated to the length of \wt{the} shortest arc in the resolved path.
$l_{upper}$ is updated to the length of \wt{the} longest arc multiplying with a coefficient $\beta\in(0,1)$.
In this way,
we obtain the path $P$ containing $t$ arcs with the same length as possible.
}

\subsection{Tours assignment}
\shs{Finally, the tours corresponding to the arcs in path $P$ are assigned to CRobs.}
To balance the workload,
a greedy strategy \cite{douglas2007modified} is employed to assign each tour to a CRob.
The tours are first sorted in descending order of length, 
and then assigned to the CRob with the least workload.


\section{Experiment and Discussion}

\subsection{Experiment Setup}
\textbf{Datasets:} 
We collect seven road networks with edge counts ranging from 914 to 12166 on Tencent Location Services (TLS) \cite{contributors2022tencent} or OpenStreetMap(OSM) \cite{contributors2017planet}.
The TLS provides grid map and we use 
\que{GVD}
to capture the graph from the grid map.
Table \ref{table:graphinfo} shows these road networks.
The first two are gained from TLS and others are gained from OSM.

\begin{table}[!t]
\caption{Information of road networks}
\begin{tabular}{|l|p{1cm}<{\centering}|p{1cm}<{\centering}|p{1cm}<{\centering}|p{1.4cm}<{\centering}|}
\hline
\multirow{2}{1.8cm}{Road networks} & No. of vertices & No. of edges & Area ($km^2$) & Network length ($km$) \\ 
\hline
BUPT         & 815  & 914   & 0.59  & 19.9  \\
Xueyuan South Rd & 3558 & 3987  & 2.56  & 85.9  \\
Yonganli    & 2289 & 2686  & 17.71 & 156.9 \\
Wangjing   & 4826 & 5651  & 31.60 & 315.1 \\
Liudaokou    & 5138 & 6124  & 47.91 & 325.2 \\
Xitucheng    & 8772 & 10381 & 68.11 & 541.3 \\
Fengtai District      & 10938 & 12166 & 116.58 & 668.3 \\
\hline
\end{tabular}
\vspace{-5pt}
\label{table:graphinfo}
\end{table}

\textbf{Implementation Details:} \label{textbf:implementationdetails}
The algorithms were implemented in C++, 
and all experiments run on an Ubuntu 20.04 laptop with an AMD R7-5800H CPU running at 3.20GHz, 16GB of RAM.
The model of CRob is Clearpath Jackal robot, 
which \que{has} a energy of \que{about} 25$km$ and a speed of \que{about} 2$m/s$.
The energy of the TRob is unlimited and its speed is set 12$m/s$,
an acceptable speed in the urban scene.
The CRob number $M$ in each team is set to $5$.
Other parameters are set as follow: 
$\alpha = 0.59$, $\tau_1 = 1000$, $\tau_2 = 100$, $\varepsilon = 1.05$, $\eta_1 = 0.02$, $\eta_2 = 0.1$, $\beta = 0.98$. 

\textbf{Evaluation Metrics:}
The metrics include computation time,
Relative Standard Deviation (RSD),
total and max tour length,
and robot utilization.
RSD represents the balance between sub-graphs and the balance between tours.
In addition,
the robot utilization is \wt{defined}: $\frac{1}{m}\sum_{i = 1}^{m}{w_i/\max(w_{1, \dots, m})}$,
where $w_{i}$ is \wt{the} tour length of $i$-th CRob,
which is proportional to working time.

\subsection{Evaluation Results}

\textbf{Effectiveness of Balanced Graph Partitioning: }
Table \ref{table:graphpartitionresult} compares our method and the K-medoids algorithm. 
We test the mean RSD and the mean runtime when a graph is split into 2 to 10 sub-graphs.
For the RSD of the sub-graph lengths,
our method outperforms the K-medoids algorithm on all graphs.
Specifically, in graph BUPT, our method performs the best, 
outperforming the K-medoids by 33.3\%.
\xf{The reason is that the clustering process obtains a preliminary balanced sub-graphs by using dynamic scale factor,
and we fine-tune the sub-graphs by re-assigning the boundary edges of sub-graphs.}
In addition,
due to the additional operation,
the computation time of our method is longer than that of K-medoids.
Since the graph partitioning is insensitive to real-time \wt{requirements},
it is appropriate to partition a very large road network with 20s time cost.

\begin{figure}[!t]
    \centering
	\includegraphics[width=1\linewidth]{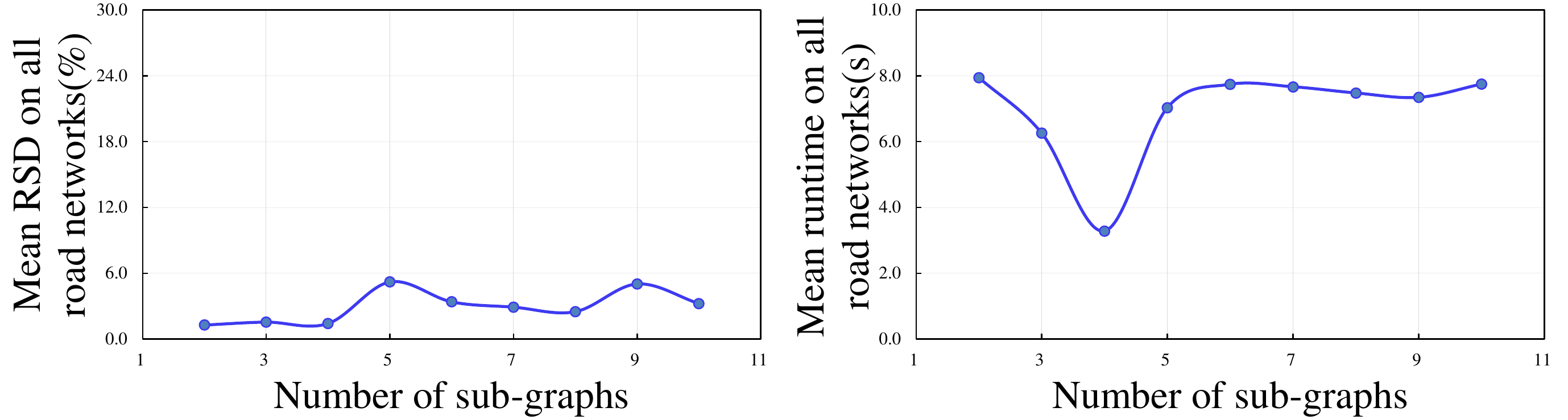}
    \caption{The mean RSD and the mean runtime of all road networks when a road network is split into 2 to 10 sub-graphs. }
    \label{fig:graphpartition}
    \vspace{-0.3cm}
\end{figure}

\xf{Fig. \ref{fig:graphpartition} shows the mean runtime and the mean of RSD for obtaining sub-graphs on multiple road networks given the number of sub-graphs.
It can be seen that our method achieves a mean RSD of less than 6\% for all given number of sub-graphs.
And the mean computation time also fluctuates less in most cases. 
It is worth noting that the computation time drops a lot when the sub-graph number is set to 4,
which is due to the complexity of \wt{the} road network.
To be specific,
the shape of the road network may affect the speed of BGP convergence in some number of clusters.
}

\begin{table}[!t]
\caption{Perfromance of graph partition on different road networks}
\vspace{-6pt}
\label{table:graphpartitionresult}
\begin{center}
\begin{tabular}{|l|cc|cc|}
\hline
\multirow{2}{1.8cm}{Road networks} & \multicolumn{2}{c|}{RSD of sub-graphs (\%)} & \multicolumn{2}{c|}{Mean Runtime ($s$)} \\
\cline{2-5} 
 & BGP & K-medoids & BGP & K-medoids \\
 \hline
BUPT         & 1.2 & 33.3 & 0.1  & 0.02 \\
Xueyuan South Rd & 1.5 & 22.1 & 1.1  & 0.42 \\
Yonganli     & 1.6 & 38.7 & 1.1  & 0.34 \\
Wangjing   & 2.0 & 58.9 & 4.7  & 1.06 \\
Liudaokou    & 1.3 & 50.8 & 5.8  & 1.82 \\
Xitucheng    & 1.7 & 41.9 & 14.7 & 6.12 \\
Fengtai District & 11.3 & 69.8 & 21.2 & 9.8 \\
\hline
\end{tabular}
\end{center}
\vspace{-3pt}
\end{table}

\begin{table*}[!ht]
\caption{Results of the algorithms on the sub-graphs of different road networks.
\ding{172}-\ding{175} represent Path Scanning (PS) \cite{BruceLGolden1983ComputationalEW}, Merge-Embed-Merge (MEM) \cite{SauravAgarwal2020LineCW}, Ulusoy Partitioning (UP) \cite{OsmanParlaktuna2009MultirobotSC}, Balanced Ulusoy Partitioning (BUP), respectively. }

\label{table:algorithmperform}
\begin{center}
\begin{tabular}{|l|c|
p{0.35cm}<{\centering}p{0.35cm}<{\centering}p{0.35cm}<{\centering}p{0.35cm}<{\centering}|
p{0.42cm}<{\centering}p{0.42cm}<{\centering}p{0.42cm}<{\centering}p{0.42cm}<{\centering}|
p{0.42cm}<{\centering}p{0.42cm}<{\centering}p{0.42cm}<{\centering}p{0.42cm}<{\centering}|
p{0.35cm}<{\centering}p{0.35cm}<{\centering}p{0.35cm}<{\centering}p{0.35cm}<{\centering}|}
\hline
\multirow{2}{*}{Road networks} &
\multirow{2}{0.95cm}{No. of subgraphs} &
\multicolumn{4}{c|}{Total runtime ($s$)} &
\multicolumn{4}{c|}{Total tour length ($km$)} &
\multicolumn{4}{c|}{Mean max tour length ($km$)} &
\multicolumn{4}{c|}{Mean RSD (\%)} \\
\cline{3-18} 
 & & \ding{172} & \ding{173} & \ding{174} & \ding{175}
 & \ding{172} & \ding{173} & \ding{174} & \ding{175}
 & \ding{172} & \ding{173} & \ding{174} & \ding{175}
 & \ding{172} & \ding{173} & \ding{174} & \ding{175} \\
\hline
BUPT                    & 1     & {\ul 0.03}    & 0.16 & \textbf{0.03} & 0.16        & 29   & {\ul 26}   & \textbf{25}   & 27            & 25.0  & {\ul 16.4}  & 24.6  & \textbf{5.8}      & 98.9  & 36.1  & \textbf{0.0}  & {\ul 7.2}  \\ 
Xueyuan South Rd       & 2     & \textbf{0.29} & 1.93 & {\ul 0.40}    & 0.98        & 137  & 129        & \textbf{117}  & {\ul 126}     & 25.0  & 24.9  & {\ul 23.5}  & \textbf{13.3}     & {\ul 16.7}  & 24.0  & 26.0 & \textbf{4.7}  \\
Yonganli                & 3     & {\ul 0.08}    & 0.43 & \textbf{0.06} & 0.18         & 247  & {\ul 235}  & \textbf{222}  & 239           & 25.0  & 24.1  & {\ul 23.5}  & \textbf{16.5}     & 42.9  & {\ul 23.7}  & 24.6 & \textbf{4.7}  \\
Wangjing                & 5     & {\ul 0.24}    & 1.24 & \textbf{0.13} & 0.44         & 492  & {\ul 460}  & \textbf{440}  & 463           & 25.0  & 24.6  & {\ul 24.0}  & \textbf{19.4}     & 24.0  & {\ul 21.0}  & 26.6 & \textbf{7.3}  \\
Liudaokou               & 5     & {\ul 0.28}    & 1.43 & \textbf{0.18} & 0.51        & 536  & 497        & \textbf{468}  & {\ul 486}     & 25.0  & 24.8  & {\ul 23.9}  & \textbf{20.7}     & 36.9  & 15.8  & \textbf{6.1}  & {\ul 9.3}  \\
Xitucheng               & 8     & {\ul 0.54}    & 2.74 & \textbf{0.40} & 0.92        & 881  & 826        & \textbf{785}  & {\ul 807}     & 25.0  & 24.5  & {\ul 23.7}  & \textbf{21.3}     & 29.6  & 27.2  & {\ul 16.7} & \textbf{5.7}  \\
Fengtai District        & 10    & {\ul 0.53}    & 2.44 & \textbf{0.29} & 0.94         & 1207 & {\ul 1078} & \textbf{1037} & 1128          & 25.0 & 24.7 & {\ul 24.2}    & \textbf{20.8}     & {\ul 18} & 19.5 & 19.5 & \textbf{8.4} \\

\hline
\end{tabular}
\end{center}
\vspace{-0.3cm}
\end{table*}

\textbf{Effectiveness of Balanced Ulusoy Partitioning: }
We test the performance of the algorithm on sub-graphs obtained by partitioning the graphs using BGP.
Our method is compared with three heuristic algorithms:
Path Scanning algorithm (PS) \cite{BruceLGolden1983ComputationalEW}, 
Merge Embed Merge algorithm (MEM) \cite{SauravAgarwal2020LineCW} and 
Ulusoy Partitioning algorithm (UP) \cite{OsmanParlaktuna2009MultirobotSC}.

\begin{figure}[!t]
	\centering
	\includegraphics[width=1\linewidth]{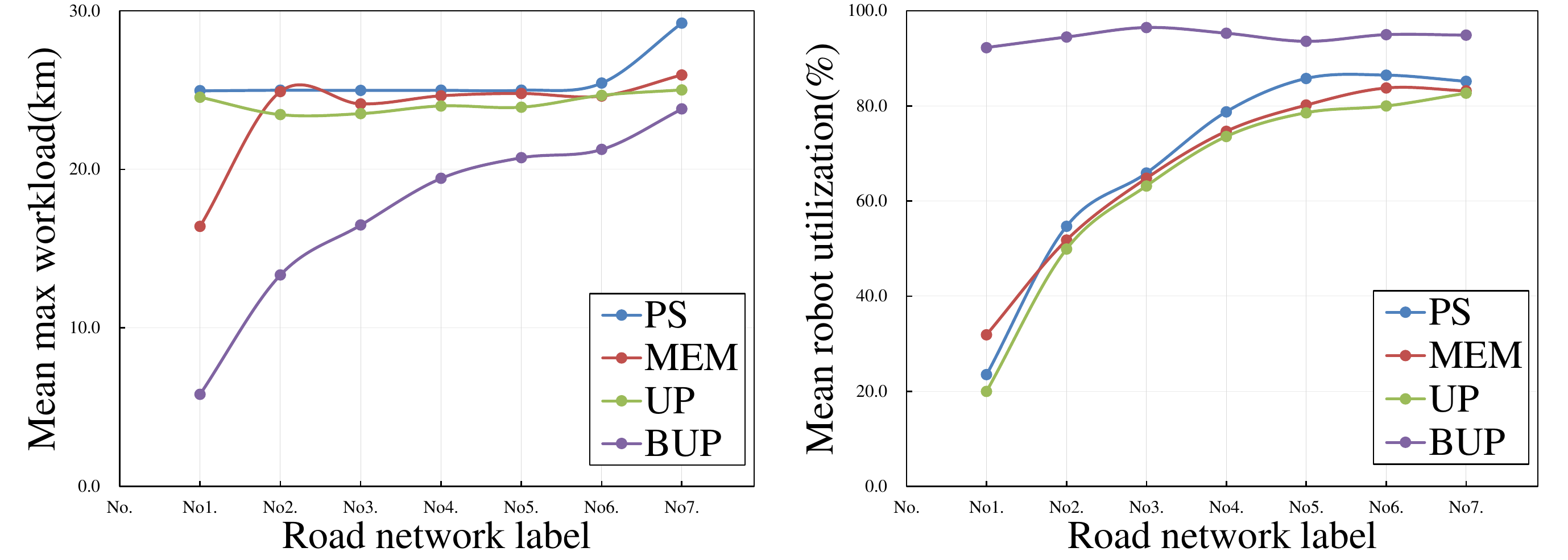}
	\vspace{-0.5cm}
	\caption{The workloads of coverage robots in different road networks.}
	\label{fig:robotworkload}
	\vspace{-0.5cm}
\end{figure}

\xf{Table \ref{table:algorithmperform} shows the total runtime and total tour length of line coverage algorithm on all sub-graphs of a given road network.
It also shows the mean of maximal tour length and the mean RSD of tour length on all sub-graphs for a given road network.}
It can be seen from \que{mean} runtime that all methods can quickly calculate the path in a sub-graph.
Although our method is not the fastest,
our speed is acceptable for line coverage.
In addition,
\xf{since our method is designed to obtain a balanced workload for robots by sacrificing a small amount of the total tour length,
BUP achieves the competitive total tour length but is not the best.}
In maximal tour length and RSD of the tour lengths,
our method achieves the best performance.
For the graph Yonganli,
our method outperforms the UP algorithm at maximal tour length with a 29.8\% reduction and achieves \wt{an} RSD of the tour lengths of 4.7\%,
\xf{which owes to the iterative process of converging arc lengths.}
\xf{Fig. \ref{fig:robotworkload} shows the \que{mean} maximal workload and the mean utilization of the CRobs on each road network.
On both metrics,
BUP outperforms others on all road networks.
In more detail,
BUP always achieves the lowest maximal workload of CRob and the highest utilization for CRob.
In addition,
the CRob utilizations of our method on all road networks are higher than 90\%,
which proves that BUP obtains \wt{a} well-balanced workload in \wt{graphs with different sizes}.}

\begin{figure}[!t]
	\centering
	\includegraphics[width=1\linewidth]{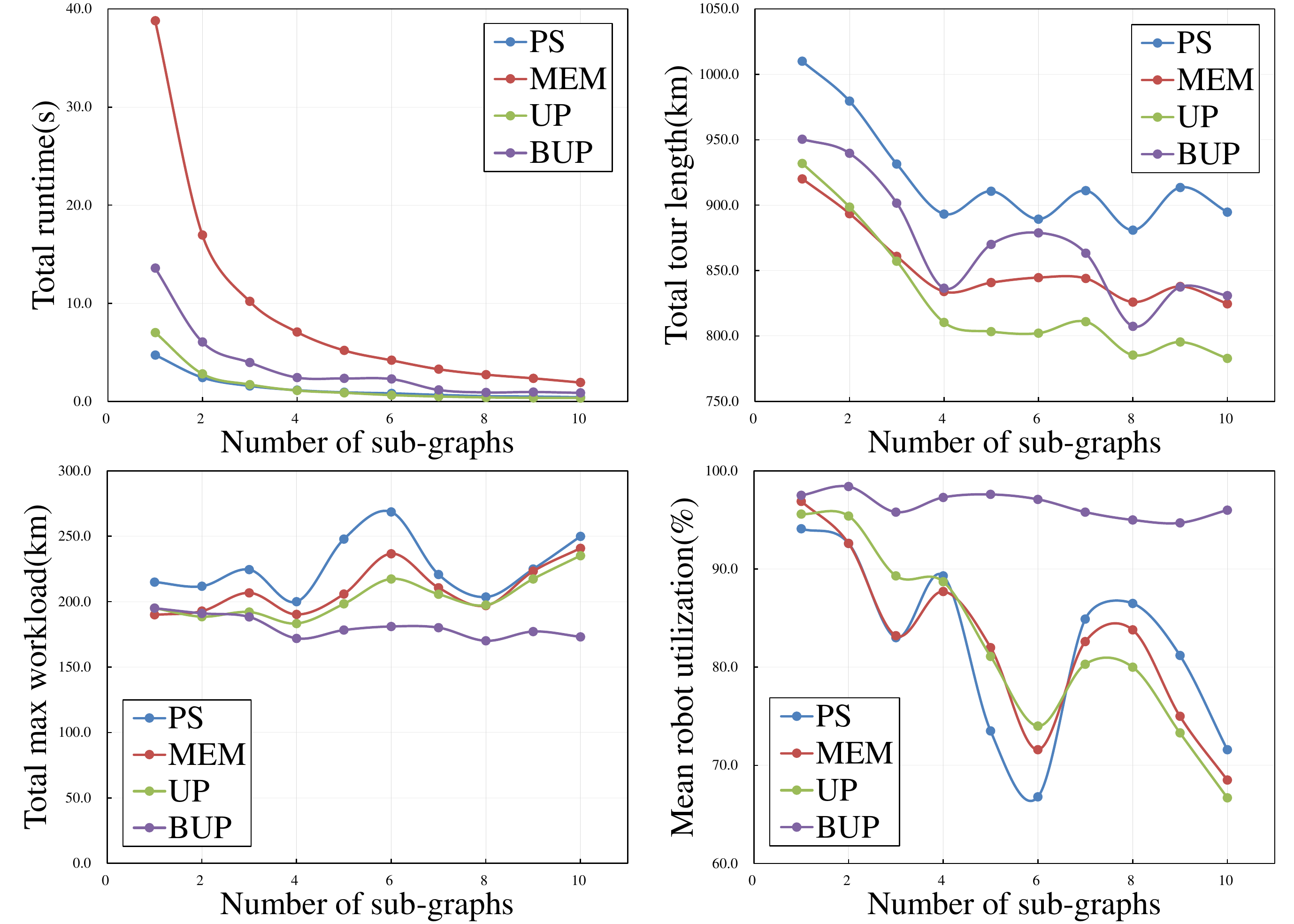}
	\vspace{-0.5cm}
	\caption{The results for different number of sub-graphs in Xitucheng.}
	\vspace{-0.2cm}
	\label{fig:linecoveragedifferentclusternum}
\end{figure}

\xf{To verify the effect of \wt{the} sub-graph number \wt{on} the line coverage performance,
Fig. \ref{fig:linecoveragedifferentclusternum} gives the results of all algorithms with the sub-graph number ranging from 1 to 10 on Xitucheng.
For the total runtime in all sub-graphs,
our method is not the fastest one,
but is competitive compared to PS and UP.
In addition,
our method achieves the best performance on \que{the} maximal workload of CRobs and robot utilization at the cost of a small increase in total tour length.}
The reason is the consideration to the relationship between the tour number and the CRob number, 
making the lengths of tours to be relatively balanced.
Thus, the CRobs have a balanced workload under different sizes of sub-graphs.
Other algorithms neglect the number of tours and the length balance of tours, 
\xf{leading to greatly performance fluctuation with the change of sub-graphs size.}

\textbf{Simulation experiment: }
\xf{To demonstrate the effectiveness of our heterogeneous approach, 
we compare \wt{the} coverage execution time of our method and \wt{a} homogeneous approach \cite{SauravAgarwal2020LineCW} on two representative large-scale road networks,
Xitucheng and Fengtai District.
For our method,
we use two teams to conduct line coverage,
each contains 1 TRob and 5 CRobs.
For another one,
we use 12 CRobs.}
Table \ref{table:simulationexperiment} shows the simulation results.
The heterogeneous system complete the task faster than the homogeneous one, 
especially in the largest road network,
task execution time is reduced by 21.7\%. 
\xf{The reason is that TRob \wt{delivers} CRob to the area for coverage,
and \wt{the} lowest max tour length makes \wt{the} lowest overall time cost.}

\begin{table}[!t]
\caption{Task execution time for different systems}
\vspace{-0.6cm}
\label{table:simulationexperiment}
\begin{center}
\begin{tabular}{|l|p{1.3cm}<{\centering}|p{1.3cm}<{\centering}|p{1cm}<{\centering}|p{1cm}<{\centering}|}
\hline
\multirow{2}{*}{Road networks} & \multicolumn{2}{c|}{Task execution time (min)} & \multirow{2}{1cm}{Time saving}&\multirow{2}{1cm}{Total length}\\
\cline{2-3}
 & Ours & MEM \cite{SauravAgarwal2020LineCW} & & \\
\hline
Xitucheng & 734.5 & 760.9 &3.5\% & 541.3$km$    \\
Fengtai District & 907.9 & 1159.4 &21.7\% & 668.3$km$   \\
\hline
\end{tabular}
\end{center}
\vspace{-0.4cm}
\end{table}

\section{Conclusion}

In this paper,
a heterogeneous multi-robot approach is proposed to achieve large-scale line coverage,
which consists of two types of robots.
In more detail, we propose a balanced graph partitioning (BGP) algorithm to balance the size of sub-graphs that needs to be covered. 
Besides,
a balanced ulusoy partitioning (BUP) algorithm is designed to obtain the balanced tours for coverage robots, 
which achieves the balanced workload of robots and reduce execution time. 
The effectiveness of our method is demonstrated by extensive experiments on various-scale road networks.

Our method set the tour count as a non-zero multiple of CRob number.
However,
we find that such multiple relationship cannot be ensured in rare experiments,
which leads to a decline in performance.
In the future,
we will delve into this issue to further improve the workload balance.


\balance
\bibliographystyle{ieeetr}
\bibliography{ref}

\end{document}